\documentclass[conference]{IEEEtran}
\IEEEoverridecommandlockouts

\usepackage[T1]{fontenc}
\usepackage[utf8]{inputenc}

\usepackage{amsmath,amssymb,amsfonts}
\usepackage{algorithmic}
\usepackage{graphicx}
\usepackage{textcomp}
\usepackage{xcolor}

\usepackage{siunitx}
\usepackage{booktabs}
\usepackage{caption}
\usepackage{hyperref}
\usepackage{tikz}

\makeatletter
\apptocmd{\@maketitle}{%
    \begin{center}
        \includegraphics[width=\linewidth]{images/teaser_introduction_comp}
        \parbox{1.0\linewidth}{
            \vspace{0.5em}
            \textbf{Physico-Geospatial Grounded Scene Interpretation.}
            By augmenting natural language scene descriptions with
            physico-geospatial knowledge like grounded buildings and path
            surfaces obtained from OpenStreetMap\cite{OSMF}, physical data like
            movement speed, relevant object counts and person
            re-identification, we ground pre-trained, unmodified VLM outputs in
            empirical reality.
        }
    \end{center}
}{}{}
\makeatother

\newcommand\submittedtext{%
	\footnotesize This work has been submitted to the IEEE for possible publication. Copyright may be transferred without notice, after which this version on arXiv will be replaced with the Version of Record.}

\newcommand\submittednotice{%
	\begin{tikzpicture}[remember picture,overlay]
		\node[anchor=south,yshift=10pt] at (current page.south) {\fbox{\parbox{\dimexpr0.65\textwidth-\fboxsep-\fboxrule\relax}{\submittedtext}}};
	\end{tikzpicture}%
}

\begin{document}
    \bstctlcite{IEEEexample:BSTcontrol}
    \title{Physico-Geospatial Grounded Scene Interpretation for Mobile Robotics\\}
    \author{
        \IEEEauthorblockN{\begin{tabular}{ccc}
                              1\textsuperscript{st} Nicolas Schuler\IEEEauthorrefmark{1}\IEEEauthorrefmark{2} &
                              2\textsuperscript{nd} Janik Kurtz\IEEEauthorrefmark{1} &
                              3\textsuperscript{rd} Lea Dewald\IEEEauthorrefmark{1}\\
                              \href{https://orcid.org/0009-0007-4098-1244}{{\includegraphics[keepaspectratio,width=0.7em]{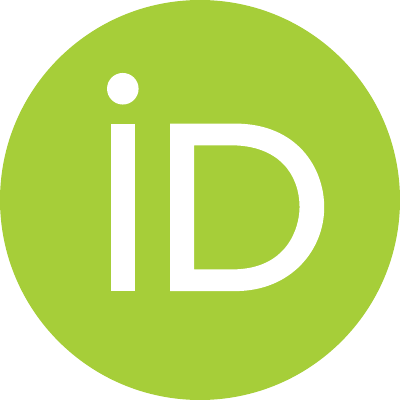}}} 0009-0007-4098-1244 &
                              \href{https://orcid.org/0009-0003-6792-2245}{{\includegraphics[keepaspectratio,width=0.7em]{images/ORCID_iD}}} 0009-0003-6792-2245 &
                              \href{https://orcid.org/0009-0004-8825-0545}{{\includegraphics[keepaspectratio,width=0.7em]{images/ORCID_iD}}} 0009-0004-8825-0545
        \end{tabular}}
        \\
        \IEEEauthorblockN{\begin{tabular}{ccc}
                              4\textsuperscript{th} Marcel Sauber\IEEEauthorrefmark{1} &
                              5\textsuperscript{th} Félicia Teferle\IEEEauthorrefmark{2} &
                              6\textsuperscript{th} Jürgen Graf\IEEEauthorrefmark{1}\\
                              \href{https://orcid.org/0009-0003-2181-0037}{{\includegraphics[keepaspectratio,width=0.7em]{images/ORCID_iD}}} 0009-0003-2181-0037 &
                              \href{https://orcid.org/0000-0002-9132-3571}{{\includegraphics[keepaspectratio,width=0.7em]{images/ORCID_iD}}} 0000-0002-9132-3571 &
                              \href{https://orcid.org/0000-0002-1354-0888}{{\includegraphics[keepaspectratio,width=0.7em]{images/ORCID_iD}}} 0000-0002-1354-0888
        \end{tabular}}
        \\
        \IEEEauthorblockA{\begin{tabular}{ccc}
                              \IEEEauthorrefmark{1} \textit{Department of Computer Science} &
                              \IEEEauthorrefmark{2} \textit{Department of Engineering}\\
                              \textit{Trier University of Applied Sciences} &
                              \textit{University of Luxembourg}\\
                              Trier, Germany &
                              Kirchberg, Luxembourg
        \end{tabular}
        }
    }

    \maketitle

    \begin{abstract}

        Recent advancements in deep learning allow robotic agents to interact
        with dynamic and unstructured environments.
        Of special interest is the integration of physico-geospatial world
        knowledge into such systems, either by using physics-aware machine
        learning models, knowledge graphs to model relationships or
        spatio-temporal and logical reasoning.
        In the present work, we introduce an approach to augment the output of
        pre-trained, unmodified VLMs used for scene interpretation by
        integrating semantic descriptions, OpenStreetMap building data and
        street information with positional, temporal and metric information
        obtained from our sensory systems, fusing this information using LLMs.
        We apply this concept to an outdoor recording within a university
        campus, achieving an F1-Score of $0.83$ in the task of grounding
        buildings and $0.64$ for path surface grounding on our pilot evaluation
        set.
        The results demonstrate the conceptual capability of the proposed
        solution to deliver physico-geospatial grounded natural language
        descriptions.
        Code and results are available at \href{https://datahub.rz.rptu.de/hstr-csrl-public/publications/physico-geospatial-grounded-scene-interpretation}{https://datahub.rz.rptu.de/hstr-csrl-public/publications/physico-geospatial-grounded-scene-interpretation}
    \end{abstract}

    \begin{IEEEkeywords}
        Mobile Robotics, Vision-Language Model, Geospatial Grounding,
        Scene Interpretation
    \end{IEEEkeywords}

	\submittednotice

    \section{Introduction}\label{sec:introduction}

    For mobile cognitive agents, interpreting their environment is crucial for
    successful interactions.
    Scene interpretation involves many sub tasks, including object detection
    and segmentation, re-identification and tracking, natural language descriptions,
    scene graph generation and reasoning.
    These challenges are often tackled by Large Language Models
    (LLMs)\cite{Bian2025,Adcock2026TheL4,Xiao2025} and Vision Language Models
    (VLMs)\cite{NVIDIA2026,Graesser,Ma2026}, enabling cognitive agents to
    describe and reason over relationships between objects and agents alike.
    The models in turn are fed by sensory systems including image and
    event cameras, LiDAR, INS/GNSS and radar\cite{Rahman2026}.
    Of particular interest is the inclusion of physical world knowledge,
    precise sensory data, uncertainty estimation and geospatial data into
    these systems.
    These physico-geospatial data allow agents to not only reason within their
    local frame of reference, but to expand their reasoning to geo-located,
    physically grounded objects\cite{Ji2025}.

    In this paper, we introduce a modular pipeline to augment
    natural language scene interpretation generated by pre-trained, unmodified
    VLMs with physico-geospatial data including speed in km/h and semantic
    enrichment, e.g.\@ the surface of a given path or the name of detected
    buildings.
    Supplementary materials including code and evaluation data have been made available
    via a \href{https://datahub.rz.rptu.de/hstr-csrl-public/publications/physico-geospatial-grounded-scene-interpretation}{public repository}.

    The remainder of this paper is structured as follows: Sect.\@
    \ref{sec:foundations-and-related-work} provides an overview of the theoretical
    background and related work.
    Next, Sect.\@ \ref{sec:automated-report-generation-pipeline}
    introduces the proposed pipeline as well as the agents and the sensory
    systems used.
    Sect.\@ \ref{sec:evaluation} discusses results given an exemplary
    application domain, and Sect.\@ \ref{sec:conclusion-and-future-work}
    concludes the paper by discussing future work.

    \section{Foundations and Related Work}\label{sec:foundations-and-related-work}

    The following section provides an overview of the theoretical foundations
    and current state-of-the-art (SOTA) models used for scene interpretation
    and geospatial reasoning.
    Sect.\@ \ref{subsec:vision-language-models} focuses on LLMs and VLMs used
    for embodied AI reasoning.
    Next, Sect.\@ \ref{subsec:geospatial-reasoning} discusses the usage of
    such architectures in the context of geospatial data in general and
    geospatial reasoning in particular.
    Sect.\@ \ref{subsec:knowledge-representation} then elucidates the
    representation of the data and knowledge retrieved by such models.
    Finally, Sect.\@ \ref{subsec:related-work} provides an overview of related
    work.

    \subsection{Vision Language Models}\label{subsec:vision-language-models}

    The transformer-based architectures of LLMs and VLMs are currently
    considered SOTA, whether for reasoning\cite{Pan2026,Zhang2026,Bjorck2026},
    scene description\cite{NVIDIA2026}, state prediction\cite{Chen2026}, as
    well as planning\cite{reuss2026state-wam,NVIDIA2026,Bjorck2026} and
    behavior tree construction\cite{Battistini2026}.
    In terms of the amount of model parameters, there are two main directions.
    On the one hand, many foundational models are closed source as well as too
    large to be deployed on edge devices\cite{Zhao2026}.
    On the other hand, there is a trend towards edge-capable models that
    can be deployed to mobile agents directly\cite{NVIDIA2026}, bypassing
    challenges like large computational requirements, bandwidth, connectivity
    and data protection / privacy concerns.

    The recently released Cosmos3\cite{NVIDIA2026} is an edge-capable world
    model for embodied AI.
    The Cosmos3 model family combines VLMs, reasoning, world and action models
    as well as generative multimodal approaches into one unified
    architecture\cite{NVIDIA2026}.
    The predecessor, the model family
    Cosmos-Reason2\cite{NVIDIA2025, NVIDIA2025a}, allows for easier
    deployment on lower computational budgets, e.g.\@ for scene descriptions.
    Similarly, Gemini Robotics ER 1.6\cite{GRT2025} focuses on physico-spatial
    visual reasoning.
    In contrast to Cosmos3 however, the model is cloud-based only, making the
    usage problematic for areas where privacy is of concern.
    For a more general overview of embodied AI and its application in robotics,
    see~\cite{Ma2026, Wang2026}.

    While such models currently focus on the sensory data received from agents
    on the ground, general-purpose image embedding models like
    DINO3\cite{Simeoni2025} can be applied to satellite imagery as well,
    leading to tighter connections of local and geospatial data, especially
    concerning geospatial reasoning.

    \subsection{Geospatial Reasoning}\label{subsec:geospatial-reasoning}

    Geospatial reasoning, i.e.,\@ problem solving based on large, diverse and
    cross-modal datasets tied to a specific location\cite{Bell2025}, has seen
    a considerable increase in interest with the rise of foundation
    models\cite{Manvi2023,Bell2025,Ji2025}.

    While general-purpose LLMs can already reason over geospatial
    data\cite{Manvi2023}, training or fine-tuning models to the specific task
    has shown beneficial results\cite{Xu2024,Wang2025a,Ji2025}.
    By incorporating geographical world knowledge into LLM architectures,
    models like EarthAI\cite{Bell2025} enable geospatial analysis via a natural
    language interface, with a particular focus on connecting graph databases
    and natural language queries\cite{Mansourian2026}.
    Similarly, GeoAgentic-RAG\cite{Liang2026} combines vector, raster, and
    geospatial data using graph databases, combining task-planning,
    spatio-temporal and visual retrieval for geospatial
    analysis\cite{Liang2026}.
    When it comes to image data, these models focus on satellite
    data\cite{Bell2025,Wang2025a} or image data from agents
    on-site\cite{Xu2024, Wu2026}.

    Open-access databases like OpenStreetMap\cite{OSMF} (OSM) or
    commercial products like Google Maps\cite{google_maps_platform_2026}
    provide information in the form of relation databases that might be used
    locally on the agent or queried on demand.

    \subsection{Knowledge Representation}\label{subsec:knowledge-representation}

    With the increased interest in geospatial databases and the need for
    integration of various types of data sources, the representation of this
    data and associated knowledge also comes into focus.
    An adequate solution must not only include geospatial data, but also be
    able to include multimodal data, often in the form of 3D point clouds, and
    its associated semantic hierarchy\cite{Tian2026}.
    In general, these approaches embed features of foundation models into
    suitable 3D representations\cite{Gorlo2025}.

    Recently, 3D scene graphs have emerged as a local, hierarchical graph
    representation combining local geometric and semantic
    information\cite{Catalano2025}, allowing for real-time reconstruction of
    world models\cite{Hughes2024}.
    While the focus in the past has largely been on indoor
    scenarios\cite{Gu2023,Werby2024},
    recent works include outdoor application as well\cite{Mukuddem2026}.
    DAAAM\cite{Gorlo2025} extends these representations to serve as
    spatio-temporal memory for LLMs queries, moving towards a more tightly
    coupled connection between reasoning and data representation.


    \subsection{Related Work}\label{subsec:related-work}

    We differentiate between local and geospatial integration of physical and
    spatio-temporal knowledge.
    That is, the first type comprises models that integrate local points of
    reference and image data from agents into their systems.
    The second type of models integrate geospatial data, including satellite
    images instead of local frames of reference only.

    For local integration, SpatialRGPT\cite{Cheng2024} integrates depth
    estimation into the embedding process to allow for spatial reasoning in
    combination with metric measurements.
    Similarly, DreamVLA\cite{Zhang2025} combines depth, dynamic and semantic
    data into a shared world embedding specifically for action planning.
    SNOW\cite{Sohn2025} encodes semantics, spatial and temporal information to
    enable grounded reasoning within a local frame of reference.

    Foundation models for geospatial reasoning like EarthAI\cite{Bell2025}
    utilize geospatial data, including satellite and population data.
    However, these models have no direct connection to an agent acting within
    a local world model, i.e.,\@ local sensory data.
    By combining VLMs and geospatial data from OSM\cite{OSMF},
    OPEN\cite{Wang2025b} allows for semantic navigation, by mapping local
    localization and planning to georeferenced objects, accessible by an
    LLM-agent for task planning.

    In contrast to such approaches, we combine local physical object properties
    like absolute speed in km/h and local object coordinates obtained by 3D
    scene reconstruction with geospatial positions of known objects like
    buildings and streets.
    To the best of our knowledge, our approach is distinct in utilizing such
    physico-geospatial data to augment scene descriptions generated by VLM
    models, grounding the output in empirical reality.

    \section{Physico-Geospatial Grounding Pipeline}\label{sec:automated-report-generation-pipeline}

    This section elucidates the proposed pipeline to generate
    physico-geospatial grounded natural language descriptions.
    First, Sect.\@ \ref{subsec:our-mobile-platforms-and-sensory-systems}
    illustrates the agents used within our laboratory.
    Then, Sect.\@ \ref{subsec:physico-geospatial-grounding-for-mobile-robotics}
    presents an overview of the pipeline, including the models used, data
    modalities, and workflow.

    \subsection{Our Mobile Platforms and Sensory Systems}\label{subsec:our-mobile-platforms-and-sensory-systems}

\begin{figure}[htbp]
        \centerline{\includegraphics[width=\linewidth]{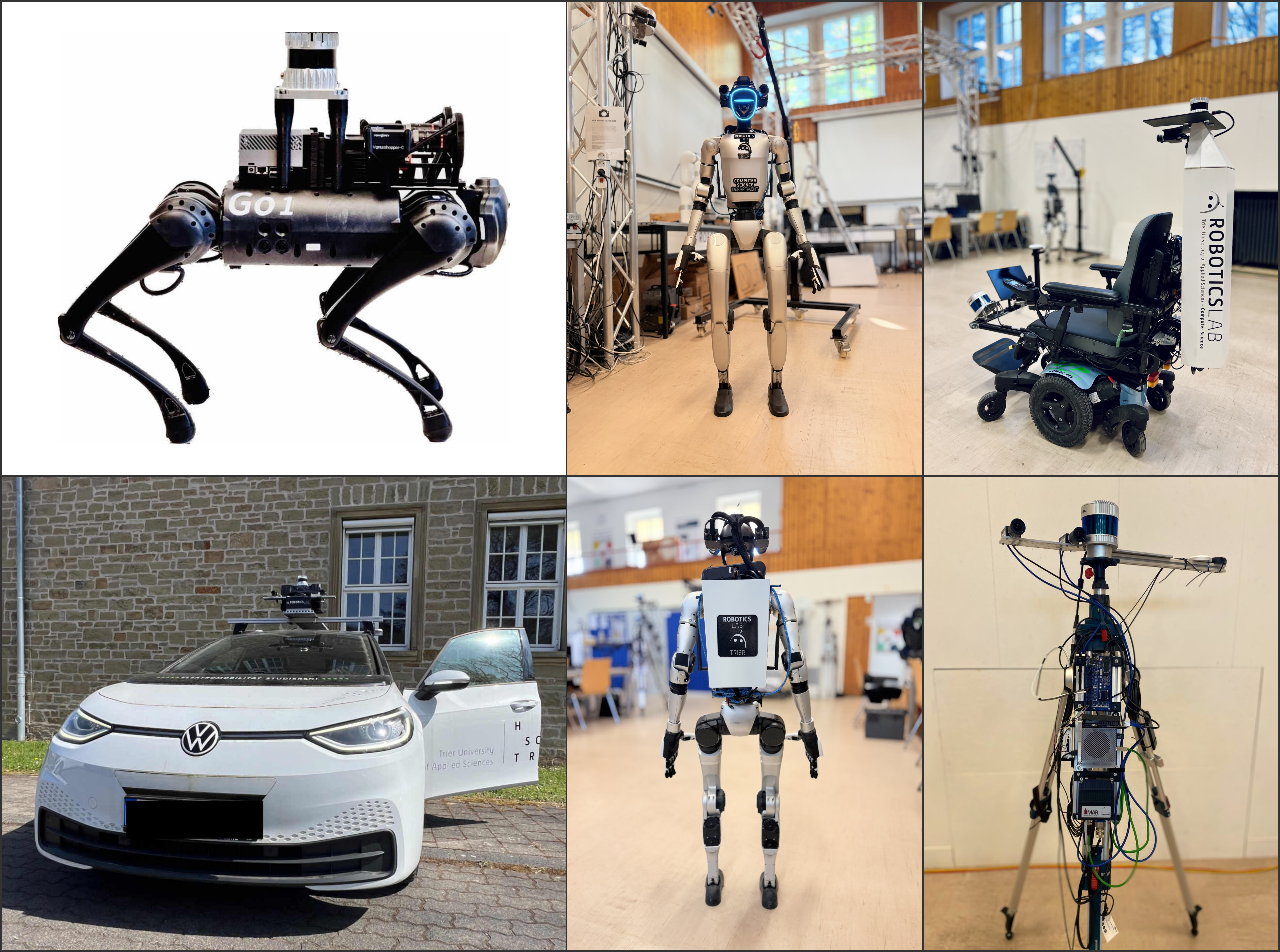}}
        \caption{
            \textbf{Different cognitive systems used within our laboratory}.
            Unitree Go1 (top left), vehicle (bottom left), Unitree G1
            (top and bottom mid), wheelchair (top right) and tripod (bottom
            right) with custom modular sensory system setups.
            Images partially taken from \cite{Schuler2025}.
        }
        \label{fig:fig1}
    \end{figure}

    The sensory data used to generate the scene descriptions and reports
    discussed below are recorded within our laboratory.
    We utilize various mobile agents and platforms, including humanoid
    (Unitree G1) and quadruped (Unitree Go1) robots, electric wheelchairs,
    vehicles and tripods (Bosch BT 300 HD), see Fig.\@ \ref{fig:fig1}.
    All of these platforms are modified by us to leverage a flexible custom
    sensory system setup including multiple video cameras
    (FLIR FG-P5G-50S4C-C), LiDAR (Ouster OS1-64 Rev.\@ 7, HESAI Pandar 64,
    LIVOX MID-360), INS (iMAR iNAT) and GNSS (u-blox F9R).
    We use an NVIDIA Jetson AGX Orin as the central computational platform
    and a custom circuit board to handle data recording.

    \subsection{Physico-Geospatial Grounding for Mobile Robotics}\label{subsec:physico-geospatial-grounding-for-mobile-robotics}

    We propose an integrated pipeline to generate physico-geospatial grounded
    scene interpretation in natural language.
    The following overview summarizes details concerning data modalities,
    models, and external databases used.

    As the first step, we calibrate our sensory setup and record the raw
    sensory data needed.
    This includes RGB stereo image data, LiDAR scans, and positional data of the
    agent using GNSS/INS\@.
    The following steps are either done after the recording is done, that is in
    a post-processing setup, or in a more limited capacity with smaller models
    live on-device.
    For the present paper, we focus on post-processing.

    In the second step, we calculate various semantic information from the raw data.
    For scene description, we utilize the model family
    Cosmos-Reason2\cite{NVIDIA2025a} to generate one-sentence natural language
    descriptions for a fixed timeframe, e.g.\ three seconds per description,
    with the input being one frame per second of one of the RGB cameras.
    A combination of Grounded DINO\cite{Liu2023} and SAM\cite{Kirillov2023} is
    used for zero-shot class segmentation.
    This allows us to flexibly detect a default set of classes as well as being
    able to dynamically detect objects relevant to the generated textual
    descriptions.
    For person re-identification, we deploy a custom,
    OSNet\cite{Zhou2019}-based tracking model, while other objects are tracked
    using ByteTrack\cite{Zhang2021}.
    The depth map associated with each image can either be estimated using the
    stereo camera data in combination with Fast-FoundationStereo\cite{Wen2025}
    or taken from the corresponding LiDAR scan.

    The third step projects the detected object masks into the world using 3D
    scene projection.
    By utilizing the recorded images, associated object segmentation masks and
    the calibrated camera matrices in combination with the estimated position
    of the agent within the world, we calculate the position for each object in
    the ECEF coordinate system.
    We either employ the centroid of the segmentation mask or the center of the
    bounding box as the point to project into the scene, depending on inference
    time constraints.

    The fourth step combines physical and geospatial grounding.
    We calculate physical properties for relevant objects, i.e.\@ the speed in
    km/h for tracked persons by applying a Kalman filter.
    In addition, we calculate the OSM\cite{OSMF} object for a given
    relevant object, e.g.\@ buildings and paths.
    To do this, we perform a lookup for a given query object, i.e.,\@ the
    coordinates of a building projected into the world, calculating the
    distances to the closest candidates.
    Using multiple linked observations over time, we estimate the best match
    for a given tracked object using a temporal Bayesian filter.
    All image-space observations are associated with OSM objects and assigned a
    track ID using ByteTrack.
    For each observation corresponding to a tracked object, we calculate the
    distance from the estimated projected world coordinates of that object to
    all relevant OSM objects in the vicinity.
    The resulting distance vector $d$ for the given observation is then
    fed into the Bayesian filter, with $\sigma = 10$ in meters for the distances to
    account for the noisy stereo depth estimation.
    The likelihood $L$ for each candidate OSM object $i$ is computed using a
    Gaussian distance model:
    $L(d_i) = \exp\left(-\frac{d_i^2}{2\sigma^2}\right)$
    The filter returns the current belief distribution over all OSM objects for
    a given observation at each timestamp.
    For each tracked object, we assign the most likely OSM object at each
    timestamp as well as the most likely object at the end of the recording.
    We utilize this final belief state as the best OSM object match for each
    track.
    Thus, after this step, each object instance for the relevant classes,
    e.g.\@ persons, buildings, paths and so on, has associated physical and
    geospatial properties linked to them.


    The fifth step uses some of this data to generate a new, grounded scene
    interpretation.
    By prompting a LLM to augment the previously generated description with the
    relevant information the natural language description is modified to
    include the physico-geospatial information retrieved above.
    For the purpose of this paper, we utilize an Open WebUI\cite{Baek2025}
    instance serving various models within a local network.

    Finally, this data can be used for various downstream tasks including
    natural language report generation\cite{Schuler2025a}.
    Within the context of this paper, we focus on the physico-geospatial
    grounded scene interpretation as a result of the pipeline only.


    \section{Evaluation}\label{sec:evaluation}

    We evaluate the feasibility of the proposed pipeline on a recording
    designed to illustrate its capabilities and limitations.
    First, we discuss the experimental setup including implementation details
    and used metrics in Sect.\@ \ref{subsec:experimental-setup}.
    Next, we present and discuss the results in
    Sect.\@ \ref{subsec:results} and Sect.\@ \ref{subsec:discussion}.

    \subsection{Experimental Setup}\label{subsec:experimental-setup}

    To evaluate the proposed pipeline, we use the baseline configuration
    described in
    Sect.\@ \ref{subsec:physico-geospatial-grounding-for-mobile-robotics}.
    The recordings were conducted on a wheelchair platform within the campus of
    Trier University of Applied Sciences, with a stereo base of 46 centimeters.
    We apply the pipeline to a continuous recording of 8464 frames at 20Hz,
    resulting in a length of seven minutes.
    The proposed pipeline is implemented using Python 3.12 and
    PyTorch\cite{Paszke2019}.
    We use an NVIDIA Jetson AGX Orin as the edge device for recording purposes
    and an NVIDIA GeForce RTX 5090 as well as AMD Ryzen 5 5600X for
    post-processing of the recorded data.
    If not specifically mentioned, we utilize the Deep Learning models
    listed in
    Sect.\@ \ref{subsec:physico-geospatial-grounding-for-mobile-robotics}.
    For tracking non-person entities, we apply ByteTrack\cite{Zhang2021},
    for object detection we use Grounded DINO\cite{Liu2023} in combination with
    SAM\cite{Kirillov2023}.

    Baseline descriptions are generated using
    Cosmos-Reason2-8B\cite{NVIDIA2025a}.
    We provide the used prompts in the \href{https://datahub.rz.rptu.de/hstr-csrl-public/publications/physico-geospatial-grounded-scene-interpretation}{supplementary repository}.
    The recording used is split into three seconds intervals, with a
    description being generated for every sequence.
    This results in 142 baseline descriptions being generated which are
    evaluated in the next section.
    For the purpose of this paper, we do not evaluate the generated baseline
    descriptions in detail and take them as-is.
    Errors introduced via the baseline descriptions, e.g.\ hallucinations,
    are not considered errors for the purpose of the evaluation.
    That is, we focus on potential errors introduced by our approach alone.

    To predict the walking speed of tracked persons, we utilize the
    calculated ECEF world coordinates projected into a local reference frame
    and use this in a Kalman filter followed by an RTS-Smoother.
    For the Kalman filter, we assume a distance standard deviation of 10 meters
    due to utilizing stereo depth reconstruction and general noise, and an
    acceleration standard deviation of 1 $\text{m/s}^2$ due to estimating the walking
    speed of persons.
    Since we do lack a ground truth for the persons' speeds, we take the value
    as-is to demonstrate the conceptual capabilities of the pipeline to
    incorporate such data without evaluating the validity of the speed values
    measured.
    The depth for a given image is calculated using our stereo camera setup and
    Fast-FoundationStereo\cite{Wen2025}, with the projected 3D point of an
    object being applied to the center of its detection bounding box as an
    evaluation-specific configuration.

    For the purpose of this paper, we calculate and provide the following
    additional information to the LLM agent for each three second sequence,
    that is baseline description:
    Identified buildings (unique name), identified paths (usage and surface
    composition), number of cars detected and name and speed of tracked
    persons.
    To integrate the generated data into the baseline description, we utilize
    Llama4:16x17b and Llama4:128x17b\cite{Adcock2026TheL4} with
    temperature $=0.0$ and 4-bit quantization.
    We provide the used prompts in the \href{https://datahub.rz.rptu.de/hstr-csrl-public/publications/physico-geospatial-grounded-scene-interpretation}{supplementary repository}.

    We evaluate the results as follows:
    For the core of the pipeline, that is the identification of buildings and
    paths with OSM objects, we calculate the F1-Score and Subset Accuracy.
    We do this for three comparisons: OSM Predictions versus Ground Truth
    (`OSM vs GT'), OSM Predictions versus Detections (`OSM vs Detection') and
    Ground Truth versus Detections (`GT vs Detection').
    The three comparisons evaluate different areas: `OSM vs GT' evaluates the
    pipeline from end to end, that is, to what extent the pipeline detects
    the OSM elements in question compared to the ground truth.
    `OSM vs Detection' evaluates the pipeline compared to the objects detected
    by the object detection models used.
    That is, if these models do not detect a specific building in question,
    the error propagates down the pipeline.
    This evaluation thus removes this source of error and focuses on the
    OSM object identification.
    Finally, `GT vs Detection' evaluates the quality of the object detection
    models used.
    Besides detecting OSM objects, we also detect and count other objects to
    enrich the augmented descriptions.
    For the present paper, we do this with cars and give the difference between
    number of detected objects within a sequence compared to the ground truth
    as the Mean Absolute Error (MAE).
    Then, we evaluate the quality of the person re-identification model used, comparing
    ground truth to detections.
    We report 95\% bootstrap confidence intervals for the Subset Accuracy
    and F1-Score respectively.

    The extracted objects are used to augment the generated baseline
    description.
    To evaluate the augmented descriptions generated by the two LLMs used
    for this task, we define eight error categories, four of which are rated by
    two independent raters.
    The four categories evaluated by raters are:
    `Hallucination' denotes errors introduced that are not based on the
    provided additional input data.
    `Wrong Entities' and `Wrong Relations' denote errors that are introduced by
    associating wrong relationships between correct objects or vice versa given
    the input information.
    Finally, `Input Description Left Out' denotes the error of leaving out
    information from the baseline description in the augmented description.
    The other four categories are not evaluated by individual raters since they
    can be automatically extracted from the generated texts.
    This includes whether the LLM completely includes the provided information,
    i.e.,\ buildings, paths, re-identified persons and speeds, as well as
    detected cars, in the augmented descriptions.
    The categories are given as error prevalence between $0.0$ and $1.0$, with
    $0.0$ denoting no error detected in any of the 142 descriptions evaluated.
    We report the Wilson 95\% confidence interval for the binary error
    classification for each subcategory.

    To measure the quality of the manual annotations of the categories by the
    raters, we calculate Cohen's $\kappa$\cite{Cohen1960} for each of the four
    rated categories for both models used, with $0.0$ denoting rating
    agreement equivalent to that expected by chance and $1.0$ total agreement
    between the raters.
    In addition, we give the `Agreement' as the ratio of agreements and total
    ratings.
    The error prevalence mentioned above is calculated on the final rating
    agreed upon by both raters by merging the two ratings and individually
    resolving any disagreement in the ratings provided by consensus.
    The original ratings, shared annotation guidelines as well as the final
    agreed upon annotations are provided in the
    \href{https://datahub.rz.rptu.de/hstr-csrl-public/publications/physico-geospatial-grounded-scene-interpretation}{supplementary repository}.

    \subsection{Results}\label{subsec:results}

    Tab.\@ \ref{tab:osm_objects} examines subtasks in the pipeline before
    using the LLM to generate the final augmented descriptions.
    In particular, we report the F1-Score and Subset Accuracy of our method to
    detect and identify buildings and paths as well as the used person re-identification\@.
    From end to end of the pipeline, that is, comparing the final found
    buildings to the ground truth, the F1-Score is $0.83$ for identified
    buildings and $0.64$ for paths.
    For our used re-identification algorithm, the F1-Score compared to the actual detections
    is $0.84$.
    Note that for persons detected, we only give values of the ground truth
    compared to the detections, since there is no intermediate step.
    For `Paths' and `Buildings', there are three points of data to consider,
    that is ground truth, object detection and object identification using
    OpenStreetMap.
    In addition, we detected cars in the recording to count the occurrence of
    these and augment the description further.
    Since we purely count the number of occurrences, metrics like Accuracy and
    F1-Score do not apply directly and we report the Mean Absolute Error
    instead, which is $3.04$ for our dataset with an average of $8.17$ and a
    median of $5$ unique cars in the ground truth per description, that is,
    three second interval.

    \renewcommand{\arraystretch}{1.2}
    \begin{table}[t]
        \centering
        \caption{\textbf{Accuracy and F1-Score of the proposed solutions}. For
        building and path identification as well as the utilized custom re-identification
        model. Values are given with 95\% confidence intervals, indicated by
        superscript and subscript notation.}
        \label{tab:osm_objects}
        \begin{tabular}{ll@{\hspace{1em}}cc}
            \toprule
            \textbf{Element} &
            \textbf{Comparison} &
            \textbf{Subset Accuracy} &
            \textbf{F1-Score} \\
            \midrule
            Buildings & OSM vs GT & $0.56^{+0.08}_{-0.08}$ & $0.83^{+0.04}_{-0.04}$
            \\
            \addlinespace
            & OSM vs Detection & $0.66^{+0.08}_{-0.08}$ & $0.87^{+0.04}_{-0.04}$
            \\
            \addlinespace
            & GT vs Detection & $0.76^{+0.06}_{-0.07}$ & $0.92^{+0.03}_{-0.03}$
            \\
            \addlinespace
            Paths & OSM vs GT & $0.40^{+0.08}_{-0.08}$ & $0.64^{+0.06}_{-0.06}$
            \\
            \addlinespace
            & OSM vs Detection & $0.53^{+0.08}_{-0.08}$ & $0.69^{+0.06}_{-0.06}$
            \\
            \addlinespace
            & GT vs Detection & $0.80^{+0.07}_{-0.07}$ & $0.90^{+0.04}_{-0.04}$
            \\
            \addlinespace
            Persons & GT vs Detection & $0.77^{+0.07}_{-0.07}$ & $0.84^{+0.05}_{-0.05}$
            \\
            \bottomrule
        \end{tabular}
    \end{table}

    Tab.\@ \ref{tab:llm_error_rates} provides the annotated errors found in
    our augmented descriptions compared to the baseline, split by the two used
    LLMs Llama4:128x17b and Llama4:16x17b.
    In general, Llama4:128x17b displays a lower error rate than Llama4:16x18b,
    with the exception of category `Input Description Left Out' with an error
    rate of $0.04$ versus $0.01$.
    For Llama4:128x17b, 30 percent of the generated descriptions contain any
    kind of error described by the given categories with 52 percent of
    descriptions for Llama4:16x17b.
    The highest error rate in any category is found for both models in
    `Wrong Entities' with $0.13$ and $0.15$ respectively.

    \begin{table}[t]
        \centering
        \caption{\textbf{LLM Error Statistic}. Error prevalence of the
        evaluated categories. Values are given with 95\% confidence intervals,
            indicated by superscript and subscript notation.}
        \label{tab:llm_error_rates}
        \begin{tabular}{l@{\hspace{1em}}cc}
            \toprule
            \textbf{Error Type} &
            \textbf{Llama4:128x17b} &
            \textbf{Llama4:16x17b} \\
            \midrule
            Hallucination & $0.00^{+0.03}_{-0.00}$ & $0.04^{+0.04}_{-0.02}$
            \\
            \addlinespace
            Wrong Entities & $0.13^{+0.07}_{-0.05}$ & $0.15^{+0.07}_{-0.05}$
            \\
            \addlinespace
            Wrong Relations & $0.02^{+0.04}_{-0.01}$ & $0.03^{+0.04}_{-0.02}$
            \\
            \addlinespace
            Input Description Left Out & $0.04^{+0.04}_{-0.02}$ & $0.01^{+0.03}_{-0.01}$
            \\
            \addlinespace
            Not Included All Objects & $0.02^{+0.04}_{-0.01}$ & $0.09^{+0.06}_{-0.04}$
            \\
            \addlinespace
            Not Included All Buildings & $0.09^{+0.06}_{-0.04}$ & $0.12^{+0.06}_{-0.04}$
            \\
            \addlinespace
            Not Included All Paths & $0.05^{+0.05}_{-0.03}$ & $0.15^{+0.07}_{-0.05}$
            \\
            \addlinespace
            Not Included All Persons & $0.07^{+0.05}_{-0.03}$ & $0.21^{+0.07}_{-0.06}$
            \\
            \addlinespace
            Any Error In Description & $0.30^{+0.08}_{-0.07}$ & $0.52^{+0.08}_{-0.08}$
            \\
            \bottomrule
        \end{tabular}
    \end{table}

    Tab.\@ \ref{tab:annotation_reliability} tabulates the annotation
    reliability as the inter-rater agreement measured by Cohen's $\kappa$ as
    well as the overall agreement in percent.
    The data is split into the two models utilized, Llama4:128x17b and
    Llama4:16x17b.
    The agreement measured by Cohen's $\kappa$ in most of the categories
    can be considered moderate ($>0.4$) to substantial ($>0.6$), with the
    highest agreement being reached in the category `Wrong Entities' for the
    model Llama4:16x17b with $\kappa=0.88$.
    For the category `Input Description Left Out' (Llama4:128x17b),
    the agreement is fair ($<0.4$).
    For the two categories `Hallucination' (Llama4:128x17b) and
    `Input Description Left Out' (Llama4:16x17b), there is agreement equivalent
    to that expected by chance between the raters.\cite{Cohen1960}

    \begin{table}[t]
        \centering
        \caption{\textbf{Annotation Reliability as inter-rater agreement}
        measured by Cohen's $\kappa$\cite{Cohen1960} and `Agreement' as a ratio.
        Annotations of augmented descriptions compared to the baseline.}
        \label{tab:annotation_reliability}
        \begin{tabular}{llS[table-format=1.2]S[table-format=3.2]}
            \toprule
            \textbf{Attribute} & \textbf{Model} &
            \textbf{Cohen's $\kappa$} & \textbf{Agreement} \\
            \midrule
            Hallucination & Llama4:128x17b & 0.000000 & 0.971831
            \\
            & Llama4:16x17b & 0.585401 & 0.97831
            \\
            Wrong Entities & Llama4:128x17b & 0.585081 & 0.880282
            \\
            & Llama4:16x17b & 0.876950 & 0.964789
            \\
            Wrong Relations & Llama4:128x17b & 0.486438 & 0.971831
            \\
            & Llama4:16x17b & 0.585401 & 0.971831
            \\
            Input Description & Llama4:128x17b & 0.320574 & 0.971831
            \\
            Left Out & Llama4:16x17b & 0.0 & 0.992958
            \\
            \bottomrule
        \end{tabular}
    \end{table}

    Besides the quantitative metrics given above, we provide a full list of
    all generated raw outputs including individual ratings, ground truths used
    to calculate the presented results as well as additional metrics for the
    entire used sequence in the
    \href{https://datahub.rz.rptu.de/hstr-csrl-public/publications/physico-geospatial-grounded-scene-interpretation}{supplementary repository}.

    \subsection{Discussion}\label{subsec:discussion}

    The following discussion is structured around three main aspects.
    First, we discuss the subsections of the proposed pipeline for building
    and path identification as well as person re-identification and object
    detection.
    Then, we discuss the integration of this data into the augmented
    description by the used LLM agents and the errors introduced by the
    process.
    Finally, we combine the previous two parts and examine the pipeline in its
    entirety given exemplary outputs.

    The results indicate that the proposed approach is in principle able to
    identify detected buildings and paths with their respective OSM objects.
    The Subset Accuracy of $0.56$ for `OSM vs GT' (`Buildings') is especially
    affected by buildings that are heavily occluded or far away, e.g., $>100m$.
    Thus, when eliminating the error introduced by the used detection models,
    the accuracy increases to $0.66$.
    Besides the detection of relevant objects, a major factor lowering accuracy
    is the depth estimation and following scene projection of the detected
    objects.
    Since we use stereo depth estimation and the detected buildings in our
    recordings can be up to 200 meters away, this introduces major noise on
    the depth maps.
    In addition, further noise is introduced if the object in question is
    obscured by objects like trees or fences, especially if these are partially
    semi-transparent and thus might be included into the object masks, heavily
    skewing the depth data.
    These challenges can be partially mitigated by using depth data from a
    LiDAR instead and better sampling of the depth maps.
    For the identification of paths and their surfaces, the challenges are
    different.
    For one, the pipeline performs noticeably worse for path predictions,
    with a Subset Accuracy of $0.40$ for `OSM vs GT'.
    We identified two reasons for this.
    First, paths are often continuous, merge into each other and might not have
    clearly defined bounds that are identifiable optically.
    This leads the generated segmentation masks to merge into adjacent
    paths, making it difficult to distinguish between them as individual
    objects for the purpose of this paper.
    While many paths might be detected by the initial models, leading to a
    higher accuracy of $0.8$ compared to buildings at $0.76$ (`GT vs
    Detection'), this does not translate to a higher accuracy when identifying
    OSM objects.
    In addition, the definition of path objects in OSM as ways with de facto no
    volume introduces errors.
    Since we perform a lookup on the closest path object to our predicted world
    coordinates, the closest path by its actual physical extension might not be
    the closest path element in OSM\@, geometrically represented as a
    centerline.
    Such is the case when walking over a parking space, with only the very
    middle containing a path object, with the edges of the parking aisle
    potentially being closer to other path objects in the vicinity.
    This problem does not occur for buildings.
    In summary, while the pipeline is able to identify relevant objects and use
    them in the following steps, the results are susceptible to erroneous
    projection of objects into the world, mostly due to noise in the depth
    estimation and object detection.


    We now focus on the integration of the generated data, i.e.,\ OSM buildings
    and paths information, relevant object count, and person re-identification
    and speed, into the baseline description using LLMs.
    The larger of the two LLMs, Llama4:128x17b, performs better than the
    smaller model, Llama4:16x17b, with an error rate of $0.3$ versus $0.52$,
    i.e., 30 percent of descriptions contain any kind of error.
    Since both models show a similar tendency, we concentrate on discussing the
    results of the larger model.
    First, the model is mostly capable of including the provided information
    into the baseline description.
    The highest error rate out of the four inclusion categories is found for
    `Not Included All Buildings' with $0.09$, that is, in nine percent of all
    descriptions not all OSM building objects provided were added to the
    augmented descriptions.
    In these cases, the model either omits the object names entirely and only
    changes the number of objects, e.g.\ `\textit{A person in dark clothing
    walks [\ldots] beside a modern building.}' is changed to `\textit{A person
    in dark clothing walks [\ldots] beside modern buildings.}',
    or the model keeps too close to the baseline structure, e.g.\ the
    baseline `\textit{A person is walking on the sidewalk beside a building
        [\ldots].}' becomes `\textit{A person is walking on the sidewalk beside
    Building K [\ldots].}' but the input includes `Building K', `Building L',
    `Building X', `Building N', `Building J'.
    These errors can potentially be reduced by improved prompting and output
    structuring.
    Looking at the other error types besides including the provided additional
    data, `Wrong Entities' is the largest contributor to erroneous augmented
    descriptions with an error prevalence of $0.13$.
    These errors come down to associating the wrong input objects with the
    baseline descriptions, e.g.\ `\textit{A person is walking on the sidewalk
    beside Building K [\ldots].}' when in reality he walks beside Building
    X\@.
    Given the input data provided, it is not possible for the model to reliably
    map these inputs to the correct relations and objects, if multiple possibly
    matching inputs are provided.
    By adding further information, like relative distance to the location of
    the agent and textual descriptions, these errors might be reduced.
    In general, augmenting the input objects with more information might
    enable the LLM to reason more precisely over the data.

    The latter discussed error categories were rated by two independent raters,
    both of which ratings were combined to a final evaluation.
    Cohen's $\kappa$ evaluates the agreement between these raters.
    For most categories, the agreement can be considered at least
    `moderate'\cite{Cohen1960}.
    However, three of the categories have a low or even chance-level Cohen's
    $\kappa$.
    We want to discuss the reasons.
    For `Hallucination' (Llama4:128x17b), the inter-rater agreement is $0.0$.
    Since for most of the rated descriptions, the default value is $0$ (no
    error found), few disagreements in ratings lowers Cohen's $\kappa$
    significantly.
    In this case, rater01 ascribed zero cases of hallucination to the outputs,
    while rater02 ascribed four cases, thus lowering the agreement rating to
    $0.0$.
    The disagreement stems from rater02 applying a more strict outlook on
    hallucinations if the semantics are open to interpretation.
    Besides the mentioned categories, the agreement is acceptable, although it
    highlights the difficulty of evaluating semantic texts that do not have a
    well-defined ground truth.

\begin{figure}[htbp]
        \centerline{\includegraphics[width=\linewidth]{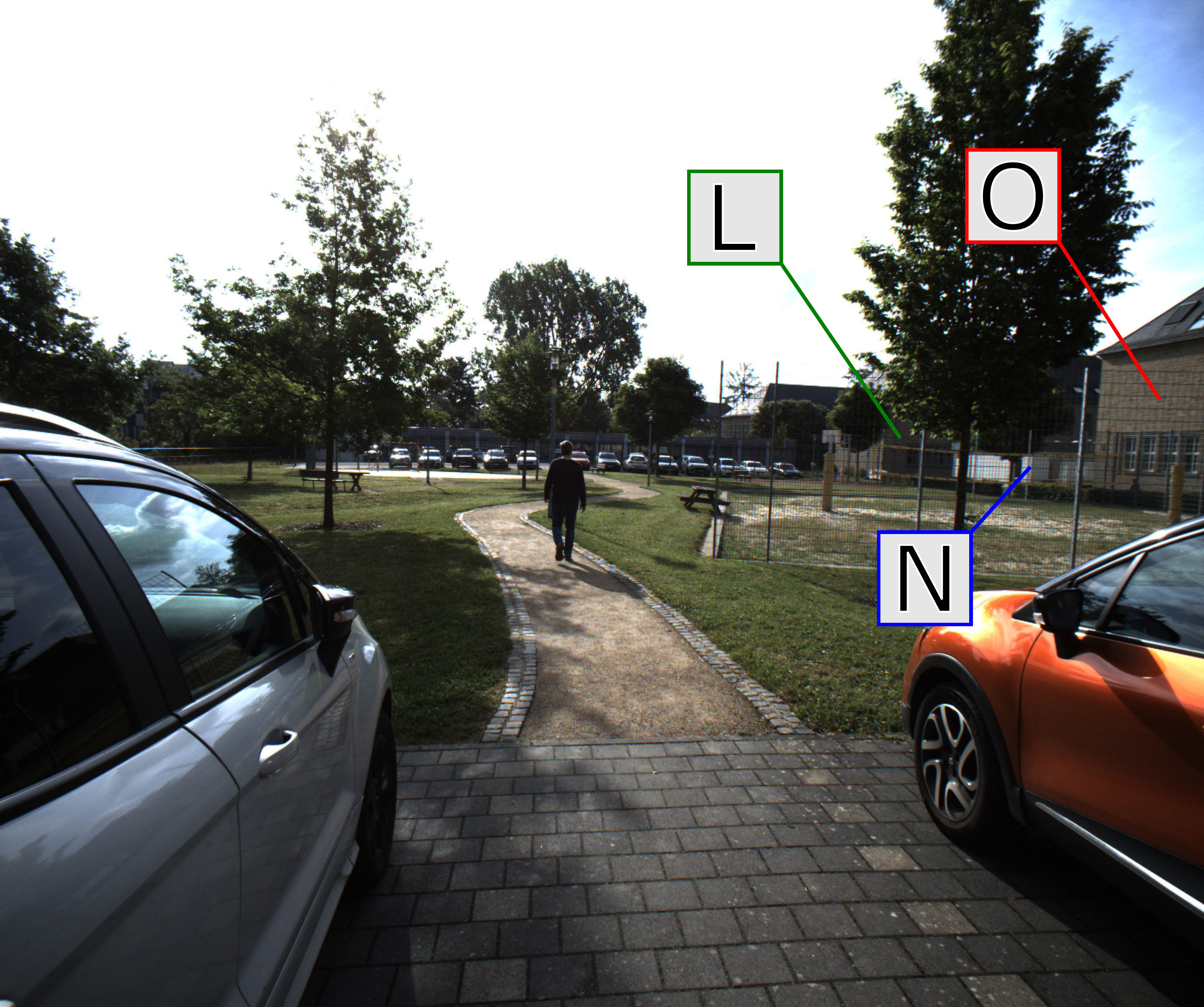}}
        \caption{
            \textbf{Exemplary image from the evaluation data}.
            Buildings O, N and L detected and identified by the pipeline,
            manually highlighted for clarity.
        }
        \label{fig:fig3}
    \end{figure}

    Combining the entire pipeline from object detection to LLM-based
    description augmentation adds empirically grounded information relative to
    the baseline.
    The generated descriptions include for the most part specific information
    regarding the correct names of buildings and surfaces of provided paths.
    If the scene is simple, i.e.,\ there are only a few, close
    buildings in view, one clearly defined path and a single person, the
    augmentation is reliable.
    This is also true if relevant information is only partially detected.
    In the following description, the pipeline does not detect all relevant
    buildings, but all provided information is correct and enhances the
    baseline without introducing errors.
    The baseline description `\textit{A person is walking along a paved path in
    a park-like setting, with cars parked on either side and trees lining the
    area.}' thus becomes `\textit{A person, Person\_01, is walking at a speed
    of 5 km/h along a sandy footpath in a park-like setting near Building O,
        with approximately 16 cars parked on either side and trees lining the
        area.}'
    While this description is still erroneous according to our rating, i.e.,
    it does not contain all provided buildings from the input, it still
    adds additional empirically grounded information to the baseline.
    The augmented description includes a correctly named building,
    identifies the person, correctly adds `sandy' to the path and specifies the
    number of relevant objects.
    In addition, two other correct buildings were detected but not included by
    the LLM in the final resulting description.
    An image from the sequence corresponding to the description with
    highlighting of the detected buildings is given in Fig.\@ \ref{fig:fig3}.

    \section{Conclusion and Future Work}\label{sec:conclusion-and-future-work}

    In the present work, we introduced a pipeline to augment natural language
    scene interpretation generated by pre-trained, unmodified VLMs by addition of
    physical measurements, e.g.\@ speed in km/h, geospatial information like
    the name of buildings encountered and the surface composition
    of paths as well as additional information in the form of detected objects
    and reidentified persons.
    We gave a demonstration of the pipeline by augmenting descriptions
    generated by Cosmos-Reason2-8B\cite{NVIDIA2025a}, discussing strengths and
    limitations of the approach.


    As preliminary tests have indicated, the addition of the augmentation data
    to the VLM input prompt directly led to no improvement of the output for
    pre-trained, unmodified VLMs.
    One direction for future work is thus fine tuning a model to the specific
    tasks to allow it to fruitfully utilize the provided data directly, without
    requiring an additional LLM\@.

    Finally, the evaluation has shown that one key challenge is the proper
    connection of the detected entities to the baseline description.
    To facilitate this, the input should be augmented, e.g.\ by adding
    relative distances of the detected objects or adding semantic descriptions
    of the identified objects.
    Further, utilizing VLMs that output grounded descriptions with
    localizations of objects contained in the environment might enable a
    stronger grounding between the baseline description and additional
    information.

    \section*{Data Availability and Protection}\label{sec:data_avail}
    The data used for this paper was recorded at the campus of Trier
    University of Applied Sciences.
    The raw data is confidential and only shared with reviewers upon
    request, with personal data being anonymized.
    All published data is anonymized and reidentified persons are
    pseudonymized.
    All recorded, confidential and published data is exclusively stored on
    servers at Trier University of Applied Sciences and the RPTU University
    Kaiserslautern-Landau.
    We make code, example video and image data, metrics and results
    available at \href{https://datahub.rz.rptu.de/hstr-csrl-public/publications/physico-geospatial-grounded-scene-interpretation}{https://datahub.rz.rptu.de/hstr-csrl-public/publications/physico-geospatial-grounded-scene-interpretation}.


    \bibliographystyle{IEEEtran}
    \bibliography{literature}

\end{document}